%% file: main.tex
\RequirePackage{fix-cm}
\documentclass[twocolumn]{svjour3}          
\smartqed  

\usepackage[colorlinks,linkcolor=blue]{hyperref}

\usepackage{cite}
\usepackage[pdftex]{graphicx}
\usepackage{ragged2e}
\usepackage[tight,footnotesize]{subfigure}
\usepackage{rotating}
\usepackage{graphicx}
\usepackage{amsmath,amssymb,amsfonts}%
\usepackage{array}
\usepackage{multirow}
\usepackage{colortbl}
\usepackage{amsfonts}
\usepackage{pifont}
\usepackage{xspace}
\usepackage{etoolbox}
\usepackage{overpic}
\usepackage{color}
\usepackage{microtype}

\usepackage{float}
\usepackage{graphicx}%
\usepackage{multirow}%

\usepackage{mathrsfs}%
\usepackage[title]{appendix}%
\usepackage{xcolor}%
\usepackage{textcomp}%
\usepackage{manyfoot}%
\usepackage{booktabs}%
\usepackage{algpseudocode}%
\usepackage{listings}%
\usepackage{blindtext}
\usepackage{graphicx} 
\usepackage{multirow} 
\usepackage{booktabs} 
\usepackage{caption}  
\usepackage{amsmath}  
\usepackage[T1]{fontenc}

\usepackage[utf8]{inputenc} 
\usepackage[T1]{fontenc}    
\usepackage{hyperref}       
\usepackage{booktabs}       
\usepackage{amsfonts}       
\usepackage{nicefrac}       
\usepackage{microtype}      
\usepackage{xcolor}         
\usepackage{graphicx}       
\usepackage{amsmath}
\usepackage{multirow}       
\usepackage{float}
\usepackage{hyperref}

\usepackage{amssymb}
\usepackage{amsmath}

\usepackage{multirow}

\usepackage{mathtools}
\usepackage[ruled]{algorithm2e}
\usepackage{amsmath}

\def\etal{\emph{et al}.}

\definecolor{RightColor}{RGB}{229,220,139}
\definecolor{WrongColor}{RGB}{234,198,189}
\definecolor{OtherColor}{RGB}{136,199,193}

\definecolor{FASColor}{RGB}{200,127,20}
\definecolor{DeepFakeColor}{RGB}{79,141,213}
\definecolor{UnifiedColor}{RGB}{254,139,0}

\newcommand{\orcid}[1]{\href{https://orcid.org/#1}{\includegraphics[width=10pt]{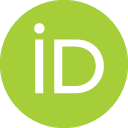}}}

\def\etal{{\em et al}}

\usepackage{hyperref}
\hypersetup{breaklinks=true,citecolor=blue, colorlinks}

\graphicspath{{./Imgs/}}
\DeclareGraphicsExtensions{.pdf,.jpg,.png}

\usepackage{silence}
\journalname{Research Article}

\begin{document}

\title{Seg-Agent: Test-Time Multimodal Reasoning for Training-Free Language-Guided Segmentation}

\titlerunning{Seg-Agent}        

\author{Chao Hao \orcid{0009-0004-4926-2543}       \and
 Jun Xu  \orcid{0000-0002-8344-2669} \and 
  Ji Du  \orcid{0009-0001-9388-8146} \and 
  Shuo Ye  \orcid{0000-0001-7756-8233} \and
  Ziyue Qiao \orcid{0000-0002-9485-4861} \and
  Xiaodong Cun \orcid{0000-0003-3607-2236} \and
  Guangcong Wang \orcid{0000-0002-6627-814X} \and
  Xubin Zheng \orcid{0000-0000-0000-0000} \and
  Zitong Yu \orcid{0000-0001-6505-3304}}

\authorrunning{C. Hao \etal} 

\institute{
Chao Hao, Shuo Ye,  Ziyue Qiao, Xiaodong Cun, Guangcong Wang, Xubin Zheng and Zitong Yu are with the School of Computing and Information Technology, Great Bay University, Dongguan, Guangdong, 523000, China. \\
Jun Xu is with Hangzhou International Innovation Institute, Beihang University, Hangzhou, 311115, China. \\
Ji Du is with the Department of Computing, The Hong Kong Polytechnic
University, Hong Kong.\\
Corresponding author: Zitong Yu (Email: yuzitong@gbu.edu.cn).
}

\date{Received: date / Accepted: date}

\maketitle

\begin{abstract}
Language-guided segmentation transcends the scope limitations of traditional semantic segmentation, enabling models to segment arbitrary target regions based on natural language instructions. Existing approaches typically adopt a two-stage framework: employing Multimodal Large Language Models (MLLMs) to interpret instructions and generate visual prompts, followed by foundational segmentation models (e.g., SAM) to produce masks. However, due to the limited spatial grounding capabilities of off-the-shelf MLLMs, these methods often rely on extensive training on large-scale datasets to achieve satisfactory accuracy. 
While recent advances have introduced reasoning mechanisms to improve performance, they predominantly operate within the textual domain, performing chain-of-thought reasoning solely based on abstract text representations without direct visual feedback. In this paper, we propose \textbf{Seg-Agent}, a completely training-free framework that pioneers \textbf{Explicit Multimodal Chain-of-Reasoning}. Unlike prior text-only reasoning, our approach constructs an interactive visual reasoning loop comprising three stages: generation, selection, and refinement. Specifically, we leverage Set-of-Mark (SoM) visual prompting to render candidate regions directly onto the image, allowing the MLLM to ``see'' and iteratively reason about spatial relationships in the visual domain rather than just the textual one. This explicit multimodal interaction enables Seg-Agent to achieve performance comparable to state-of-the-art training-based methods without any parameter updates. Furthermore, to comprehensively evaluate generalization across diverse scenarios, we introduce \textbf{Various-LangSeg}, a novel benchmark covering explicit semantic, generic object, and reasoning-guided segmentation tasks. Extensive experiments demonstrate the effectiveness and robustness of our method. Our code and dataset will be available at \href{https://github.com/Fanye12/Seg-Agent}{https://github.com/Fanye12/Seg-Agent}.
\end{abstract}

\section{Introduction}
\label{sec:intro}

The rapid development of multimodal large language models (MLLMs) \cite{llava, gpt4, qwenvl2.5} and foundational segmentation models \cite{maskformer, mask2former, SAM, SAM2} has driven significant progress in language-guided segmentation \cite{groundedsam, lisa, segllm, text4seg}. Unlike traditional segmentation methods \cite{segformer, SETR, GRES, SENet, yuv20k, JoNet}, which are limited to predefined categories and scenarios, language-guided segmentation models can segment any target region of interest based on textual instruction. This makes it an open and domain-unrestricted segmentation approach.

\begin{figure*}[t]
\centering
\includegraphics[width=0.8\linewidth]{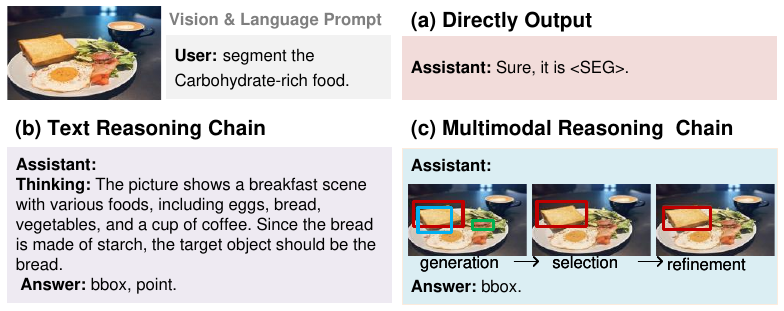}
\caption{Comparison of different reasoning paradigms for language-guided segmentation, . (a) Directly Output: Training-based methods where MLLM directly generates visual prompt without reasoning. (b) Text-only Chain-of-Reasoning: Training-based methods where MLLM performs reasoning in the textual domain before outputting coordinates. (c) Our Seg-Agent: We construct an explicit multimodal reasoning chain (generation, selection, refinement) where the model reasons directly in the visual domain through Set-of-Mark prompting, enabling iterative self-correction at test-time without any training.}
\label{fig:2}
\end{figure*}

Most existing language-guided segmentation models follow a two-stage approach \cite{lisa, pixellm, sam4mllm, segzero}. First, they use MLLMs to understand the instruction and perceive the image, generating visual prompts (typically in the form of bounding boxes or points). Then, a foundational segmentation model such as SAM is employed to produce high-quality segmentation masks based on these visual prompts. However, due to limitations such as the MLLM's relatively weak spatial perception and grounding capabilities, the visual prompts it generates directly are often of low quality \cite{lisa, SoM}. 

\begin{figure*}[t]
\centering
\includegraphics[width=0.9\linewidth]{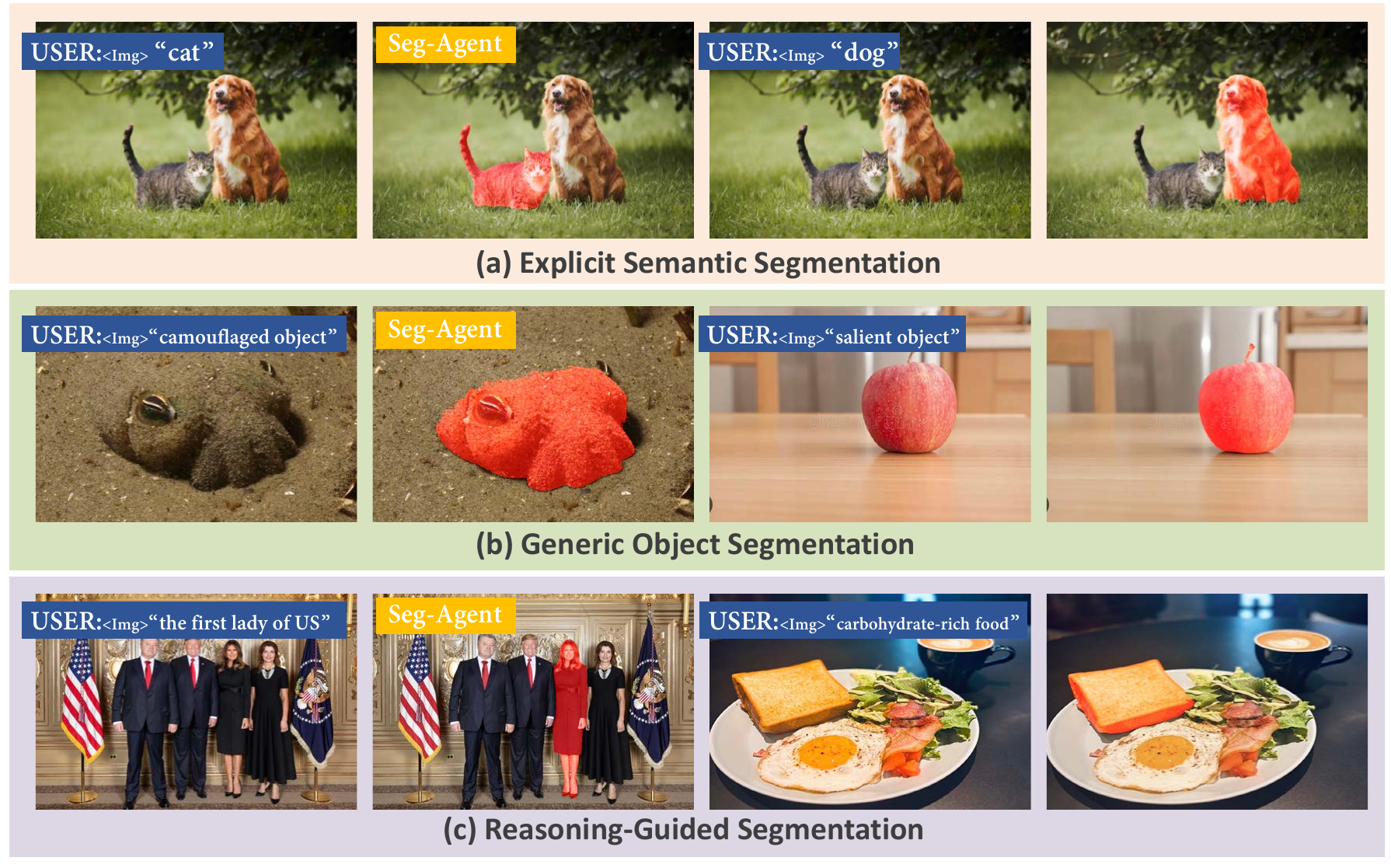}
\caption{Given an image and a textual target, Seg-Agent can handle segmentation tasks across various scenarios: (a) \textbf{Explicit Semantic Segmentation}: segmenting objects with clearly defined semantics (e.g., ``cat'', ``dog''). (b) \textbf{Generic Object Segmentation}: segmenting conceptually defined objects without specific categories (e.g., ``camouflaged object'', ``salient object''). (c) \textbf{Reasoning-Guided Segmentation}: segmenting targets based on prompts that require commonsense or factual reasoning (e.g., ``the first lady of US'', ``carbohydrate-rich food'').}
\label{fig:1}
\end{figure*}

To address this issue, existing methods typically rely on large-scale supervised training to improve the quality of generated prompts. As illustrated in Figure~\ref{fig:2}, these training-based approaches generally fall into two categories: (1) Directly Output, where the trained MLLM directly outputs visual prompts without explicit reasoning (e.g., LISA \cite{lisa}); and (2) Text-only Reasoning Chain, where the model performs reasoning purely in the textual domain before generating coordinates (e.g., Seg-Zero \cite{segzero}). While training improves performance, it introduces significant limitations: it requires collecting extensive datasets, consumes substantial computational resources, and cannot instantly leverage newer, more powerful base models. Moreover, we argue that for spatial tasks like segmentation, \textbf{text-only reasoning is insufficient} because it lacks direct interaction with visual evidence. The model essentially guesses coordinates based on abstract semantic features rather than verifying them against the image content.

To overcome these limitations, we propose \textbf{Seg-Agent}, a novel framework that shifts the paradigm from training-based alignment to test-time multimodal reasoning. As shown in Figure~\ref{fig:2}(c), instead of relying on parameter updates or text-only reasoning, we construct an \textbf{explicit multimodal chain-of-reasoning} comprising three interactive stages: generation, selection, and refinement. Specifically, we leverage Set-of-Mark (SoM) visual prompting \cite{SoM} to render candidate bounding boxes directly onto the image, allowing the MLLM to \textbf{``see'' and reason about} its own predictions in the visual domain. Through this multimodal reasoning process, the model can compare candidates visually, identify misalignments, and iteratively refine the prompts based on explicit visual feedback. This closed-loop reasoning mimics human cognitive behavior, hypothesizing, verifying, and correcting, achieving high-precision grounding without any training. By conducting this reasoning at test-time, Seg-Agent remains completely \textbf{training-free}, offering a low-cost and future-proof solution.

Moreover, existing language-guided segmentation datasets \cite{lisa, refcoco, refcocog} primarily focus on narrow referring expressions, lacking the diversity needed to comprehensively evaluate model generalization. To bridge this gap, we construct a multi-scenario evaluation benchmark called \textbf{Various-LangSeg}. As categorized in Figure~\ref{fig:1}, we divide language-guided segmentation tasks into three challenging types: Explicit Semantic Segmentation (ESS) with clearly defined categories (e.g., ``cat'', ``dog''), Generic Object Segmentation (GOS) guided by general concepts (e.g., ``camouflaged object'', ``salient object''), and Reasoning-Guided Segmentation (RGS) requiring abstract reasoning (e.g., ``the first lady of US'', ``carbohydrate-rich food''). Specifically, Various-LangSeg encompasses the three task types illustrated in Figure~\ref{fig:1} (ESS, GOS, and RGS), collectively covering the majority of common language-guided segmentation scenarios. We evaluate the performance of Seg-Agent and several related language-guided segmentation models \cite{lisa, sam4mllm, segzero} on Various-LangSeg to validate their robustness and generalization capability.

We summarize our contributions as follows:
\begin{itemize}
    \item We propose Seg-Agent, a novel framework that introduces an explicit multimodal chain-of-reasoning for language-guided segmentation. By leveraging visual feedback loops at test-time, Seg-Agent achieves performance comparable to training-based methods while remaining completely training-free.
    \item We collect a comprehensive evaluation dataset named Various-LangSeg, which covers most common scenarios of language-guided segmentation and effectively assesses models' generalization ability.
    \item Extensive experiments demonstrate the effectiveness of our proposed method and provide a low-cost, simple, and effective design paradigm for the community.
\end{itemize}

\section{Related Work}
\label{sec:relatedwork}

\subsection{Multimodal Large Language Models}
In recent years, MLLMs have achieved revolutionary progress in vision-language tasks. Models such as GPT-4 \cite{gpt4} and Qwen2.5-VL \cite{qwenvl2.5} have demonstrated outstanding capabilities in understanding multimodal content, giving them a natural advantage in tasks requiring joint image-text reasoning, thus driving the development of numerous downstream tasks \cite{androidworld, guigrounding}. MLLMs have shown remarkable performance in visual question answering \cite{vqa}, image captioning \cite{captioning}, and multimodal reasoning \cite{zhang2024multimodal}. However, they typically lack fine-grained spatial perception and grounding abilities \cite{visionllmv2, lisa}, which poses challenges for dense prediction tasks such as segmentation. Previous approaches mostly enhance grounding capabilities by training on task-specific data \cite{omg-llava, pixellm}. In contrast, our work leverages native MLLMs to generate visual prompts, but circumvents their limitations in spatial understanding through an explicit multimodal reasoning chain.

\subsection{Foundational Segmentation Models}
Foundational segmentation models such as MaskFormer \cite{maskformer, mask2former} and SAM \cite{SAM, SAM2} are general-purpose models trained on large-scale data for universal segmentation. In particular, SAM introduces a promptable interface that enables segmentation with sparse visual cues like points or bounding boxes. These models offer strong generalization ability and high-quality mask prediction across various domains. In our framework, we use SAM2 as the backend segmentation module and focus on improving its performance by enhancing the quality of the visual prompts generated by MLLMs.

\subsection{Language-Guided Segmentation}
Early methods typically employ a text encoder \cite{bert, clip} to extract textual features for guiding the segmentation model \cite{groundedsam, groundingdino, LAVT}. With the advancement of MLLMs and foundational segmentation models, most existing approaches  have evolved into the two-stage framework described earlier, although they differ in subtle aspects. PixelLLM \cite{pixellm} and OMG-LLaVA \cite{omg-llava} use their own lightweight decoders to generate masks. However, such decoders generally underperform compared to SAM, which is pretrained on massive-scale data. As a result, current mainstream methods directly integrate SAM as the segmentation backbone. LISA \cite{lisa}, Sa2VA \cite{sa2va} and GSVA \cite{GSVA} do not generate explicit visual prompts, instead, they introduce a special \textit{$<$SEG$>$} token to compress textual information, requiring fine-tuning of SAM so that it can interpret this novel type of prompt. In contrast, SAM4MLLM \cite{sam4mllm} and Seg-Zero \cite{segzero} adopt a more intuitive strategy: they keep SAM frozen and instead post-train MLLMs to generate more accurate bounding boxes or points prompts, formats that SAM can directly understand, thereby achieving improved performance. These two works are the closest to our Seg-Agent, except that we enhance the quality of generated visual prompts through a manually designed, explicit reasoning chain without any training.

\section{Various-LangSeg: A Comprehensive Evaluation Benchmark}
\label{sec:dataset}

\subsection{Motivation}
Existing language-guided segmentation datasets primarily focus on referring segmentation \cite{refcoco, refcocog} and reasoning segmentation \cite{lisa, mmr}. However, these datasets cover a limited spectrum of task types, which hinders the thorough evaluation of general-purpose models. To address this gap, we introduce \textbf{Various-LangSeg}, a unified and diverse benchmark designed to evaluate language-guided segmentation methods across a broad range of scenarios.

\subsection{Task Categorization}
As shown in Figure~\ref{fig:1}, we categorize language-guided segmentation tasks into three representative scenarios: 

\begin{itemize}
    \item \textbf{Explicit Semantic Segmentation}: The target is clearly specified by a direct category name (e.g., ``dog").
    \item \textbf{Generic Object Segmentation}: Tasks are category-agnostic and guided by general concepts (e.g., ``salient object or camouflaged object'').
    \item \textbf{Reasoning-Guided Segmentation}: The textual query requires abstract reasoning to infer the target (e.g., ``carbohydrate-rich food").
\end{itemize}
These three scenarios jointly span most common language-guided segmentation tasks, enabling comprehensive evaluation of model generalization.

\subsection{Dataset Construction}
Given an input image and a textual instruction, the goal is to generate the corresponding binary mask. Since our method does not require any training, Various-LangSeg is designed solely for evaluation purposes.

To construct the dataset, we proceed as follows:

1. Image Collection: We sample images and corresponding masks from existing public datasets \cite{NC4K, PASCAL-S, SBU, casia, coco}.

2. Instruction Annotation: For each image-mask pair $(x_{\text{img}}, y_{\text{mask}})$, we manually annotate a textual instruction $x_{\text{txt}}$ to form the complete triplet $(x_{\text{img}}, x_{\text{txt}}, y_{\text{mask}})$.


Note that although the images and their corresponding masks are sourced from other datasets, we must manually analyze the relationship between each image and its corresponding mask, assign accurate textual instructions, and carefully select high-quality image-mask pairs. This process is not merely a simple sampling and assembly to form a new dataset.

For each scenario:
\begin{itemize}
    \item \textbf{Explicit Semantic Segmentation}: 20 common object categories (e.g., cat, dog, bird) with 7 images per category, totaling 140 samples.
    \item \textbf{Generic Object Segmentation}: 4 binary segmentation tasks: salient object detection (SOD) \cite{sod}, camouflaged object detection (COD) \cite{cod}, shadow detection (SD) \cite{SBU}, and image tampering detection (ITD) \cite{casia}. Each task includes 16 samples. Textual input directly uses the task name (e.g., ``salient object").
    \item \textbf{Reasoning-Guided Segmentation}: 40 samples are annotated using complex descriptions requiring implicit reasoning. These are inspired by ReasonSeg \cite{lisa}, such as ``carbohydrate-rich food", where identification requires reasoning and domain knowledge.
\end{itemize}

 In total, Various-LangSeg contains $244$ evaluation samples, with $140$, $64$, and $40$ samples in three scenarios respectively. Its scale is comparable to that of the validation set of the ReasonSeg dataset \cite{lisa}. 
 We visually present the statistics of Various-LangSeg in Figure 5 in the Appendix.
We prioritize diverse scenario coverage over sheer sample size to effectively stress-test model generalization.
 
\subsection{Evaluation Metrics}
We follow established works \cite{lisa, segzero} and adopt two standard metrics: 
\begin{itemize}
    \item \textbf{gIoU (global IoU)}: the average IoU over all samples.
    \item \textbf{cIoU (cumulative IoU)}: the ratio of the total intersection area to the total union area across the entire dataset.
\end{itemize}

\begin{figure*}[ht]
\centering
  \includegraphics[width=\linewidth]{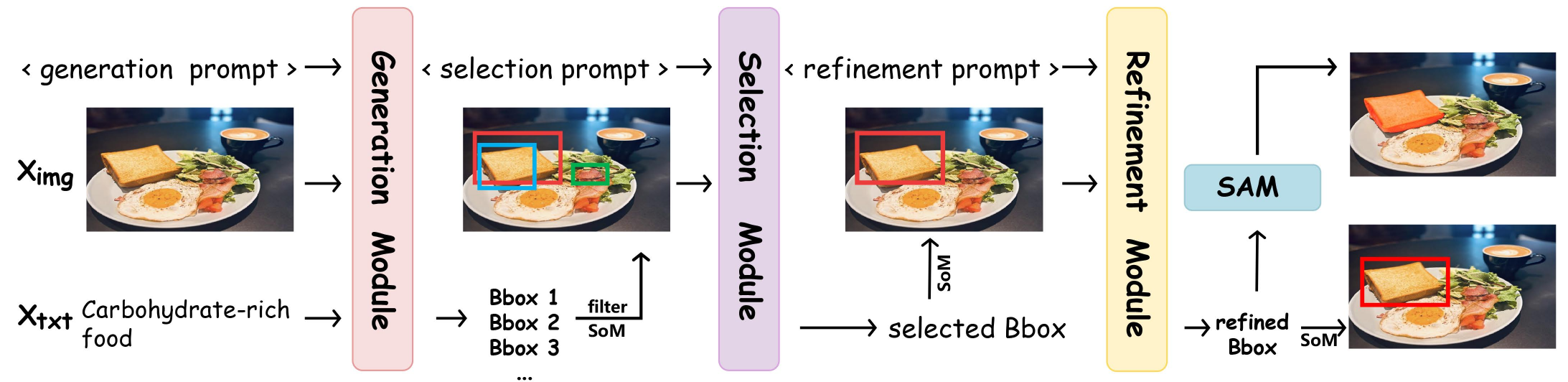}
      \caption{Illustration of Seg-Agent. By constructing an explicit multimodal reasoning chain: generation, selection, and refinement, the MLLM is able to improve the quality of generated visual prompts, thereby enabling SAM to produce more accurate target segmentation masks. SoM here indicates Set-of-Mark prompt \cite{SoM}.
}
\label{fig:3}
\end{figure*}

\section{Method}
\label{sec:method}

We propose Seg-Agent, a training-free framework for language-guided segmentation. Our method is a two-stage approach that first uses MLLMs to generate visual prompts and then employs a base segmentation model to produce the final mask. However, unlike previous training-based methods, we construct a test-time multimodal reasoning chain to guide the MLLM's prompt generation process, comprising generation, selection, and refinement, mimicking an iterative procedure of progressively localizing and fine-tuning the target boundaries. In contrast to traditional end-to-end approaches that rely on learned features, Seg-Agent explicitly constructs and updates visual prompts, guiding the segmentation model through interpretable, step-by-step interactions. The overall pipeline is illustrated in Figure~\ref{fig:3}.

\subsection{Problem Formulation}
Let $\mathbf{X}_{\text{img}} \in \mathbb{R}^{H \times W \times 3}$ be an input image and $\mathbf{X}_{\text{txt}} \in \mathcal{T}$ be a natural language instruction describing the target object (e.g., ``the man in red''). The goal is to predict a binary mask $\mathbf{Y}_{\text{mask}} \in \{0,1\}^{H \times W}$ that segments the described object.

\subsection{Overview of Seg-Agent Pipeline}

Seg-Agent consists of four modules:
\begin{enumerate}
    \item \textbf{Generation Module}: Proposes diverse bounding boxes using image augmentations.
    \item \textbf{Selection Module}: Selects the most appropriate box via visual comparison.
    \item \textbf{Refinement Module}: Fine-tunes the selected box to better align with the object boundaries.
    \item \textbf{Segmentation}: Applies a pretrained segmenter such as SAM to produce the final mask.
\end{enumerate}

\subsection{The Forward Pass of Seg-Agent}

\textbf{Generation Module.} To ensure robustness across views and scales, we apply a set of augmentations $\mathcal{A} = \{a_i\}_{i=1}^N$ to the input image (including flipping, scaling, etc):
\begin{equation}
    \mathbf{X}^{(i)} = a_i(\mathbf{X}_{\text{img}}), \quad i = 1, \dots, N.
\end{equation}
Each augmented image $\mathbf{X}^{(i)}$ is paired with $\mathbf{X}_{\text{txt}}$ and sent to an MLLM with a task-specific prompt (denoted as $<$generation prompt$>$) to localize the object:
\begin{equation}
\mathbf{B}^{(i)} = \text{MLLM}_{\text{gen}}(\mathbf{X}^{(i)}, \mathbf{X}_{\text{txt}}, \texttt{<generation prompt>}).
\end{equation}
This yields a set of bounding box proposals (coordinate format: $[x_1, y_1, x_2, y_2]$):
\begin{equation}
    \mathcal{B}_{\text{gen}} = \left\{ \mathbf{B}^{(i)} \right\}_{i=1}^N.
\end{equation}
\textbf{Selection Module.} To consolidate candidate boxes back into the original image frame, we invert the augmentations:
\begin{equation}
    \tilde{\mathbf{B}}^{(i)} = a_i^{-1}(\mathbf{B}^{(i)}).
\end{equation}
We perform Non-Maximum Suppression (NMS) to filter redundant boxes:
\begin{equation}
    \mathcal{B}_{\text{sel}} = \text{NMS} \left( \left\{ \tilde{\mathbf{B}}^{(i)} \right\}, \theta_{\text{IoU}} \right).
\end{equation}
Using a visualization strategy such as Set-of-Mark (SoM) \cite{SoM}, we render the candidate boxes onto the original image. This method has been shown to enhance the spatial perception and grounding capability of MLLMs \cite{SoM, androidworld}, and also allows for intuitive visualization of the spatial relationship between the bounding box and the target object, as illustrated in Figure~\ref{fig:3}. The MLLM receives the SoM-marked image, textual instruction, and a comparison prompt (denoted as $<$selection prompt$>$) to choose the most relevant box:
\begin{equation}
\begin{split}
    \mathbf{B}_{\text{sel}} = \text{MLLM}_{\text{sel}}(&\text{SoM}(\mathbf{X}_{\text{img}}, \mathcal{B}_{\text{sel}}),
    \mathbf{X}_{\text{txt}}, \mathcal{B}_{\text{sel}}, \\ &\texttt{<selection prompt>}).
\end{split}
\end{equation}
\textbf{Refinement Module.} The selected box $\mathbf{B}_{\text{sel}}$ may still require fine-tuning for optimal spatial coverage. We invoke a final reasoning step using another refinement prompt (denoted as $<$refinement prompt$>$) to refine it:
\begin{equation}
\begin{split}
    \mathbf{B}_{\text{refined}} = \text{MLLM}_{\text{refine}}(&\text{SoM}(\mathbf{X}_{\text{img}}, \mathbf{B}_{\text{sel}}), 
    \mathbf{X}_{\text{txt}}, \mathbf{B}_{\text{sel}},\\ &\texttt{<refinement prompt>}).
\end{split}
\end{equation}
Through this reasoning process, the MLLM is able to carefully examine the alignment between the current bounding box and the target object, and fine-tune it based on semantic and visual context, for example, by expanding, shrinking, translating, or adjusting its boundaries to achieve more precise coverage of the target region.

The final output $\mathbf{B}_{\text{refined}}$ is a well-refined bounding box that serves as a high-quality visual prompt for the subsequent segmentation task. This module significantly improves the accuracy of boundary localization and is a key step toward achieving high-precision segmentation.


\noindent\textbf{Segmentation Module.} The refined bounding box is then used as the visual prompt input to a segmentation model such as SAM:
\begin{equation}
    \mathbf{Y}_{\text{mask}} = \text{SAM}(\mathbf{X}_{\text{img}}, \mathbf{B}_{\text{refined}}).
\end{equation}
SAM uses the bounding box as a spatial prompt to precisely identify and segment the described target region in the image, producing a high-quality, pixel-level binary mask $\mathbf{Y}_{\text{mask}}$, where 1 indicates the target region and 0 indicates the background.

This step completes the final transformation from a language instruction to an accurate segmentation, serving as the last stage of the entire Seg-Agent framework. Thanks to the progressive refinement of visual prompts in the previous three stages, SAM receives more accurate guidance, thereby significantly improving the final segmentation accuracy.

Note that we provide the complete versions of the prompts introduced above in the supplementary materials.

\subsection{Why Seg-Agent Matters}
Seg-Agent decomposes language-guided segmentation into interpretable sub-tasks, enabling robust performance across diverse conditions without any task-specific training. By explicitly engaging in step-wise reasoning, Seg-Agent avoids common failure modes of end-to-end systems and provides traceable decision-making, all while fully leveraging the generalization power of MLLMs.

Unlike previous works relying on parameter updates, Seg-Agent operates in a zero-shot and training-free setting, relying solely on step-wise reasoning within MLLMs. This has several advantages:
\begin{itemize}
    \item \textbf{Generalization:} Augmentation-enriched proposals increase robustness across unseen distributions. We perform no post-training on MLLMs, thus avoiding any potential negative impacts on their performance.
    \item \textbf{Interpretability:} Each reasoning step is explicit and traceable, enabling transparent debugging and user intervention.
    \item \textbf{Modularity:} Seg-Agent can be instantly adapted to newer and stronger MLLMs or segmentation models without retraining.
\end{itemize}

\section{Experiments}
\subsection{Experimental Setting}

\textbf{Implementation Details.}
We employ QwenVL-2.5 \cite{qwenvl2.5} as the base MLLM for generating visual prompt, the generation module, selection module, and refinement module are all built upon it, which can be deployed locally or accessed via API services. Additionally, we use SAM2-Large \cite{SAM2} to generate segmentation masks. Seg-Agent operates in a training-free manner, and the entire inference process can be completed on a single NVIDIA RTX 4090 GPU with 24 GB of memory. For the NMS step, we set the IoU threshold to $0.8$. During inference, the prompt for Seg-Agent is pre-defined (see Appendix), requiring only the input of the target object. For other methods, we set the text input according to the templates provided in their works.

\input{table/ref}

\noindent\textbf{Datasets.} 
Since our method does not require training, we only select benchmark datasets to evaluate model performance. Following prior related work \cite{lisa, sam4mllm, segzero}, we adopt three datasets for the referring segmentation task: refCOCO, refCOCO+ \cite{refcoco}, and refCOCOg \cite{refcocog}. These datasets involve simple textual descriptions such as “the man wearing white clothes”, belonging to the explicit semantic segmentation scenario. We also include the ReasonSeg \cite{lisa} dataset for the reasoning segmentation task, which falls under the reasoning-guided segmentation scenario and contains referring expressions that require reasoning, such as “the food with the most Vitamin C”. Finally, we introduce the Various-LangSeg dataset proposed in this paper, which covers three common scenarios in language-guided segmentation and effectively evaluates the model's generalization and versatility.

\noindent\textbf{Baseline Methods.}
We compare Seg-Agent with two groups of methods: (1) Training-based methods, including traditional methods that do not use MLLMs such as groundedsam \cite{groundedsam}, ReLA \cite{GRES} etc., and MLLM-based methods such as LISA \cite{lisa}, SAM4MLLM \cite{sam4mllm}, etc.; (2) Training-free methods, we mainly adopt the Qwen2.5-VL + SAM2 \cite{qwenvl2.5, SAM2} baseline method introduced in Seg-Zero \cite{segzero}, with related settings and prompts kept consistent with that paper, i.e., letting MLLMs directly output bounding box coordinates in a single-step method to prompt SAM, rather than having an explicit thinking process like Seg-Agent.

\input{table/reasoning}
\noindent\textbf{Evaluation Metrics.} We follow prior work \cite{lisa, segzero} in adopting two evaluation metrics: gIoU and cIoU, which have been described before. Since cIoU tends to be heavily biased toward large objects and exhibits high variability, gIoU is generally preferred as the primary metric.

\input{table/various}
\begin{figure*}[t]
      \centering
      \includegraphics[width=0.95\linewidth]{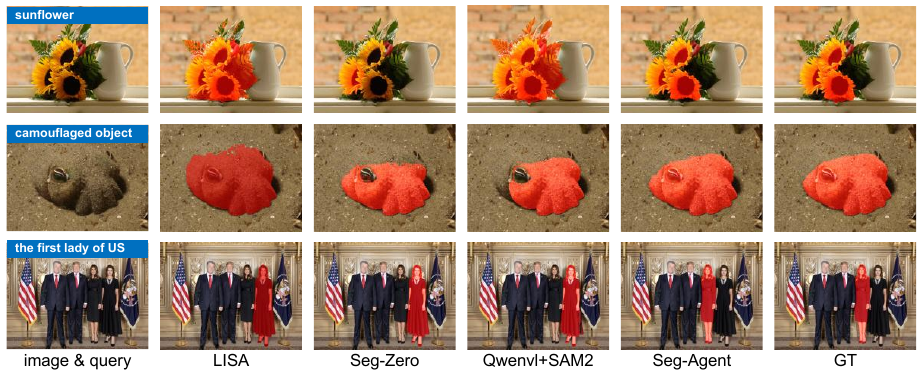}
      \caption{Visual comparison between Seg-Agent and related methods. We show three common scenarios of language-guided segmentation. }
      \label{fig:4}
\end{figure*}

\begin{figure*}[htbp]
      \centering
\includegraphics[width=0.85\linewidth]{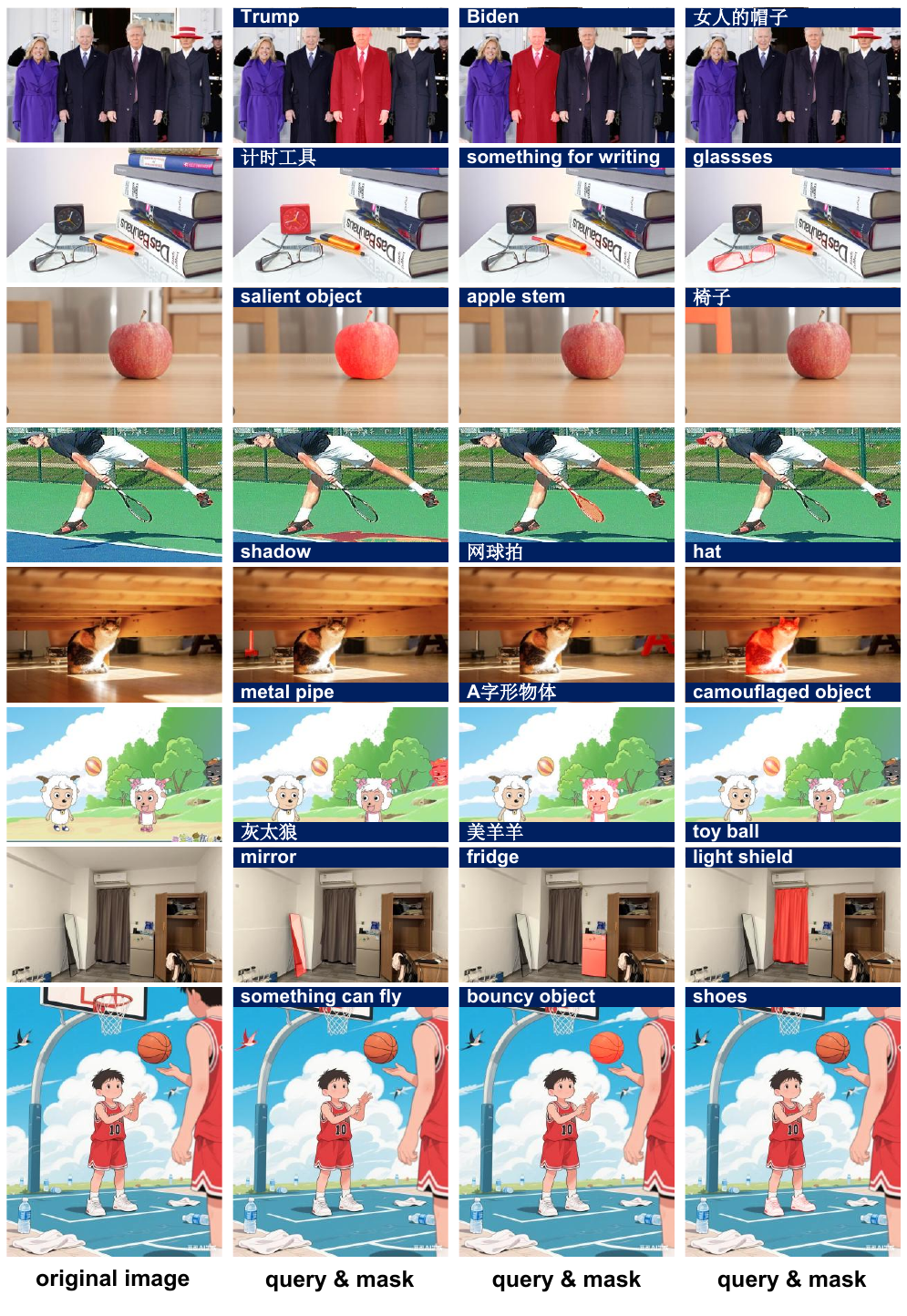}
      \caption{More visualization results of Seg-Agent. Seg-Agent can handle language inputs in various forms, including both Chinese and English. It is also capable of processing different types of images, such as real-world photos, captured photographs, cartoon images, and AI-generated images. These images demonstrate Seg-Agent's strong generalization ability and its broad range of application scenarios. Please zoom in for a better view.}
      \label{fig:more}
      \vspace{-5mm}
\end{figure*}
\subsection{Comparison With Other Methods}
We conduct a comparative analysis of the performance between Seg-Agent and several most relevant methods. We compare CRIS \cite{cris}, LAVT \cite{LAVT}, OVSeg \cite{ovseg}, X-Decoder \cite{xdecoder}, SEEM \cite{SEEM}, ReLA \cite{GRES}, LISA \cite{lisa}, PixelLLM \cite{pixellm}, PerceptionGPT \cite{PerceptionGPT}, SAM4MLLM \cite{sam4mllm} and Seg-Zero \cite{segzero}.

\vspace{1mm}
\noindent\textbf{Referring Segmentation.}  We evaluate Seg-Agent and other related methods on the test sets of refCOCO, refCOCO+, and refCOCOg in Table~\ref{tab:ref}. As mentioned earlier, the target objects in these datasets are described using simple and direct text descriptions, which falls under the explicit semantic segmentation scenario. Due to the relatively simple nature of this task, all methods achieve good performance, and there is little difference between traditional methods and MLLM-based methods. Notably, the baseline method Qwen2.5VL+SAM2-L also achieves favorable results, while Seg-Agent provides a certain improvement over this baseline and achieves performance comparable to the SOTA training-based method Seg-Zero \cite{segzero}.

\vspace{1mm}
 \noindent\textbf{Reasoning Segmentation.} 
As shown in Table~\ref{tab:reasoning}, we evaluate Seg-Agent and other related methods  on the validation and test sets of ReasonSeg \cite{lisa}. As previously mentioned, the target objects in this dataset are described using reasoning-based textual expressions, categorized into long queries and short queries, falling under the reasoning-guided segmentation scenario. Compared to explicit semantic segmentation, this task requires reasoning to first identify the target object, making it significantly more challenging. As can be seen from the table, traditional methods that do not employ MLLMs perform poorly in this scenario, as their text encoders are typically better at extracting textual features but lack reasoning capabilities. In contrast, MLLM-based methods achieve substantial performance improvements. Moreover, the training-free baseline method Qwen2.5VL+SAM2-L also achieves favorable performance, while Seg-Zero achieves significant gains over this baseline through GRPO \cite{grpo} training. Similarly, our Seg-Agent achieves notable improvement over the same baseline and attains performance comparable to the SOTA training-based method Seg-Zero, surpassing earlier approaches such as LISA.

\vspace{1mm}
 \noindent\textbf{Various-LangSeg.} As shown in Table~\ref{tab:various}, we evaluate Seg-Agent and related methods on Various-LangSeg. 
It can be observed that traditional methods perform poorly on this dataset, achieving acceptable results only in the explicit semantic segmentation scenario, and even then, they are outperformed by MLLM-based approaches. Their performance is also weak in the general object segmentation scenario and particularly poor in the reasoning-guided segmentation scenario.  In contrast, MLLM-based methods are applicable across all three scenarios. However, due to the lack of training on relevant data and the absence of an explicit reasoning process, early methods such as LISA perform relatively poorly on the general object segmentation task, underperforming compared to their results on the other two tasks.
Seg-Zero achieves the best performance among training-based methods, especially in the reasoning-guided scenario, benefiting from task-specific training.
For training-free methods, Seg-Agent consistently improves upon the Qwen2.5-VL+SAM2-L baseline in all three scenarios. Notably, Seg-Agent-7B achieves the best overall performance among training-free methods, and even outperforms many training-based approaches, highlighting its strong generalization and reasoning ability without task-specific training.

\vspace{1mm}
\noindent \textbf{Visual Comparison.}
Figure~\ref{fig:4} presents a visual comparison of Seg-Agent with several of the most relevant methods, including LISA, Seg-Zero, and Qwen2.5-VL-7B+SAM2-L. In the first row (explicit semantic segmentation scenario), the target object is a sunflower, other methods either over-segment or under-segment the object, while only Seg-Agent produces an accurate and precise segmentation. In the second row (general object segmentation scenario), other methods fail to fully capture the main structure or produce overly coarse boundaries, whereas Seg-Agent successfully preserves the overall structure without edge blurring. In the final row (reasoning-guided segmentation scenario), only Seg-Agent correctly identifies the first lady of US (second from the right), while all other methods select incorrect objects. These visual results demonstrate the strong generalization capability and segmentation accuracy of Seg-Agent. We present more visualization results in the supplementary materials.

We provide more visualization results in Figure~\ref{fig:more}. As can be seen, Seg-Agent is capable of handling inputs in various forms. It supports multilingual inputs, with a primary demonstration of Chinese, English, and their mixed usage. Furthermore, we present a variety of image types, including images from web news, datasets, screenshots, cartoons, photographs taken by cameras, and AI-generated images. The examples also cover the three types of language prompts introduced in the main text: explicit semantic segmentation, generic object segmentation, and reasoning-guided segmentation.
These high-quality segmentation results demonstrate the strong generalization capability and broad applicability of Seg-Agent. 

\input{table/ablation}

\subsection{Ablation Study}
\textbf{Effectiveness of Each Module}. 
As summarized in Table~\ref{tab:ablation}, integrating the Generation Module (GM) with either the Selection Module (SM) or Refinement Module (RM) consistently outperforms the baseline, while the full three-stage pipeline achieves optimal performance across all scenarios. This progressive gain validates the efficacy of our explicitly constructed multimodal reasoning chain: each stage addresses a distinct bottleneck in visual grounding, and their synergy enables iterative self-correction that single-step prompting cannot achieve.
Notably, the magnitude of improvement varies across task types. In the reasoning-guided segmentation (RGS) scenario, where queries demand multi-hop inference (e.g., ``carbohydrate-rich food''), the modular reasoning chain yields the most substantial gains (+5.1 gIoU over baseline). This aligns with our hypothesis that complex semantic reasoning benefits most from visual verification: by rendering candidate boxes via Set-of-Mark prompting, the MLLM can ground abstract concepts against concrete visual evidence, reducing hallucination in coordinate prediction.
All results are obtained under a fixed backbone configuration (Qwen2.5-VL-7B + SAM2-L) to isolate the contribution of our reasoning framework. The first row of Table~\ref{tab:ablation} corresponds to the single-step baseline (no GM/SM/RM), while the GM+RM configuration demonstrates that even without explicit selection, iterative refinement can partially compensate for initial proposal errors, though at the cost of generating fewer candidate hypotheses.


\input{table/setting}

\noindent\textbf{Generalization across Base Models}. We conduct experiments to demonstrate that Seg-Agent can be directly adapted to different base models in Table~\ref{tab:setting}. Under the same configuration, Seg-Agent achieves performance improvements compared to direct inference. Moreover, it can be observed that Seg-Agent is compatible with different MLLMs \cite{qwenvl2.5, internvl3} and segmentation models \cite{SAM, SAM2}, and the stronger the base model, the better Seg-Agent performs. This highlights the advantage of our proposed training-free approach: as newer and more powerful base models emerge, Seg-Agent can directly integrate with them to achieve improved segmentation performance, which is a significant advantage over training-based methods.

\subsection{Analysis and Limitations}
\label{sec:analysis_limitations}

Compared to baseline methods, Seg-Agent introduces a multi-step reasoning process, requiring multiple MLLM calls during inference. This inevitably incurs additional latency over single-step approaches. 
In contrast, most competitive baselines, such as LISA\cite{lisa} and Seg-Zero\cite{segzero}, rely on extensive training on large-scale datasets, often involving costly fine-tuning of the MLLM. These methods consume significant computational resources and are tightly coupled to their training data, limiting generalization and adaptability to new foundation models. Our approach shifts the resource burden from training to inference: instead of improving visual prompts via training-time optimization, Seg-Agent enhances them through test-time reasoning. This design aligns with our goal of a completely training-free framework. The modest latency increase is offset by dramatically lower entry barriers and greater flexibility. However, although Seg-Agent makes multiple MLLM calls (generation, selection, refinement), the total output token count is comparable to text chain-of-reasoning methods \cite{segzero, thinkbeforesegment} like Seg-Zero, which require lengthy textual reasoning 
during inference. Thus, our approach incurs no significant 
disadvantage in overall inference cost compared to other 
CoT-based segmentation methods.

Moreover, the effectiveness of Seg-Agent stems from its explicit instruction to the MLLM on how to reason, which is essentially the same in spirit as methods \cite{deepseekr1} that enhance model performance by teaching trained models how to think. This structured guidance enables the model to better align language instructions with visual content, even without parameter updates. Furthermore, Seg-Agent’s performance is inherently dependent on the capabilities of the underlying MLLM and segmentation model. However, this dependency is a feature rather than a flaw: because our method is training-free, it can seamlessly leverage any advances in foundation models. As newer, more powerful MLLMs or SAM variants become available, Seg-Agent can integrate them immediately, without any architectural changes or retraining. In practice, just 3–5 additional MLLM calls per sample can significantly boost segmentation accuracy, offering a lightweight yet effective path toward state-of-the-art performance.

Finally, we acknowledge that Various-LangSeg (244 
samples) is modest in size, comparable to the ReasonSeg 
validation set ( 200 samples), which motivated our initial 
collection scale. Our primary goal was not scale but cover-age: it is the first evaluation benchmark that unifies explicit, 
generic, and reasoning-guided segmentation scenarios in 
a single, coherent testbed. This comprehensive design enables holistic assessment of generalization across diverse 
language–guided segmentation. We plan to release an expanded, larger-scale V2 version in the future.

\section{Conclusion}
In this paper, we propose Seg-Agent, a completely training-free language-guided segmentation model. By constructing an explicit test-time multimodal reasoning chain of generation, selection, and refinement to guide the model in generating more accurate visual prompts, Seg-Agent achieves segmentation performance comparable to training-based methods. Additionally, we construct the Various-LangSeg dataset containing multiple scenarios, which can comprehensively evaluate the generalization capability of language-guided segmentation models. Extensive experiments demonstrate the effectiveness of our approach. We hope that our simple yet effective method can provide rich inspiration to the community.

\begin{small}

\vspace{.3in} \noindent \textbf{Abbreviations:}
LGS: language-guided segmentation; ESS:
Explicit Semantic Segmentation; 
GOS: generic object segmentation;
RGS: reasoning-guided segmentation.

\vspace{.3in} \noindent \textbf{Authors' Contributions:}
Chao Hao proposed the main idea of the paper, conducted most of the experiments, and wrote the manuscript.
Jun Xu, Ji Du and Shuo Ye constructed the Various-LangSeg dataset.
Ziyue Qiao, Xiaodong Cun, Guangcong Wang and Xubin Zheng participated in the discussion of the paper writing and helped polish and revise the article.
Zitong Yu supervised the project, organized the research discussions,
and finalized the writing. All authors read and approved
the final manuscript.

\vspace{.3in} \noindent \textbf{Fundings:}
This work was supported by the National Natural Science Foundation of China (Grant No. 62576076). The
computational resources are supported by SongShan
Lake HPC Center (SSL-HPC) in Great Bay University.

\vspace{.3in} \noindent \textbf{Data Availability Statement:}
All the datasets used in this study are publicly available at \href{https://github.com/lartpang/awesome-segmentation-saliency-dataset}{https://github.com/lartpang} and \href{https://github.com/JIA-Lab-research/Seg-Zero}{https://github.com/JIA-Lab-research/Seg-Zero}.

\vspace{.3in} \noindent \textbf{Competing Interests:}
The authors declare that they have no competing interests.



\end{small}

\vspace{-0.5mm}
\bibliographystyle{unsrt}
\bibliography{reference}

\end{document}

%% file: table/ref.tex
\begin{table}[t]
\centering
\caption{Referring segmentation results. Methods marked with “*” are traditional approaches, while the other methods are based on MLLMs. We compare cIoU in this table. Best results are highlighted in \textbf{bold}.}
\resizebox{\linewidth}{!}{
\begin{tabular}{l|ccc}
\toprule
Method & \multicolumn{1}{c}{RefCOCO} & \multicolumn{1}{c}{RefCOCO+} & \multicolumn{1}{c}{RefCOCOg} \\
                  & testA & testA  &  test \\
\midrule
\multicolumn{4}{c}{\small training-based methods}   \\ \hline
CRIS*               & 73.2  & 68.1 &  60.4 \\
LAVT*              & 75.8  & 68.4 &  62.1 \\
ReLA*               & 76.5  & 71.0 &  66.0 \\
LISA-7B           & 76.5  & 67.4 &  68.5 \\
PixelLM-7B         & 76.5  & 71.7 &  70.5 \\
PerceptionGPT-7B   & 78.6  & 73.9 &  71.7 \\

Seg-Zero-3B  & 79.3  &  73.7& 71.5\\ 
Seg-Zero-7B & \textbf{80.3} & \textbf{76.2} &  \textbf{72.6}\\  \hline
\multicolumn{4}{c}{\small training-free methods}   \\ \hline
Qwen2.5-VL-3B + SAM2-L &75.9   &71.5  &70.1 \\
Qwen2.5-VL-7B + SAM2-L &77.8   &73.5  &71.2 \\
Seg-Agent-3B (Ours)  &79.0   &73.2  &71.4 \\ 

Seg-Agent-7B (Ours) & \textbf{79.9} & \textbf{76.0} & \textbf{72.2} \\ 

\bottomrule
\end{tabular}}
\label{tab:ref}
\end{table}

%% file: table/reasoning.tex
\begin{table}[t]
\centering
\footnotesize
\caption{Reasoning segmentation results. 
“-” indicates the results are not available.}
\resizebox{\linewidth}{!}{
\begin{tabular}{ccccc}
\toprule
\multicolumn{1}{l|}{\multirow{3}{*}{Method}} & \multicolumn{4}{c}{ReasonSeg}                                                            \\ \cline{2-5} 
\multicolumn{1}{c|}{}                        & \multicolumn{2}{c|}{Val}                              & \multicolumn{2}{c}{Test}         \\ \cline{2-5} 
\multicolumn{1}{c|}{}                        & \multicolumn{1}{c}{gIoU} & \multicolumn{1}{c|}{cIoU} & \multicolumn{1}{c}{gIoU} & cIoU \\ \hline
\multicolumn{5}{c}{\small training-based methods}                                                                                                        \\ \hline
\multicolumn{1}{l|}{X-Decoder*}                    &            22.6               & \multicolumn{1}{c|}{17.9}     &  21.7                         & 16.3     \\
\multicolumn{1}{l|}{SEEM*}                    &            25.5               & \multicolumn{1}{c|}{21.2}     &  24.3                         & 18.7     \\
\multicolumn{1}{l|}{ReLA*}                    &            22.4               & \multicolumn{1}{c|}{19.9}     &  21.3                         & 22.0     \\
\multicolumn{1}{l|}{OVSeg*}                   &   28.5                        & \multicolumn{1}{c|}{18.6}     &    26.1                       &  20.8    \\ 
\multicolumn{1}{l|}{Grounded-SAM*}                    &            26.0              & \multicolumn{1}{c|}{14.5}     &  21.3                         & 16.4    \\
\multicolumn{1}{l|}{LISA-7B-LLaVA1.5}                    &            53.6              & \multicolumn{1}{c|}{52.3}     &  48.7                        & 48.8     \\
\multicolumn{1}{l|}{LISA-13B-LLaVA1.5}                    &            57.7              & \multicolumn{1}{c|}{60.3}     &  53.8                        & 50.8     \\
\multicolumn{1}{l|}{SAM4MLLM}                    &            46.7              & \multicolumn{1}{c|}{48.1}     &  -                         & -     \\
\multicolumn{1}{l|}{Seg-Zero-3B}                    &            58.2              & \multicolumn{1}{c|}{53.1}     &  56.1                         & 48.6     \\
\multicolumn{1}{l|}{Seg-Zero-7B}                    &           \textbf{62.6}              & \multicolumn{1}{c|}{\textbf{62.0}}     &  \textbf{57.5}                         & \textbf{52.0}     \\
\hline
\multicolumn{5}{c}{\small training-free methods}                                                                                                       \\ \hline
\multicolumn{1}{l|}{Qwen2.5VL-3B+SAM2-L}       &  53.6                         &    \multicolumn{1}{c|}{44.0}     &   47.9                        &  37.8    \\
\multicolumn{1}{l|}{Qwen2.5VL-7B+SAM2-L}                &   57.6                        & \multicolumn{1}{c|}{48.3}     & 50.1                          & 41.2     \\
\multicolumn{1}{l|}{Seg-Agent-3B (Ours)}               &  57.8                         & \multicolumn{1}{c|}{56.1}     &  55.5                         & 49.8     \\ 
\multicolumn{1}{l|}{Seg-Agent-7B (Ours)}               & \textbf{61.7}                          & \multicolumn{1}{c|}{\textbf{61.2}}     &    \textbf{57.6}                       & \textbf{51.8}     \\
\bottomrule
\end{tabular}}

\label{tab:reasoning}
\end{table}

%% file: table/various.tex
\begin{table*}[htbp]
\centering
\caption{Results on Various-LangSeg. We report the performance across three scenarios and the overall performance.}
\resizebox{0.95\linewidth}{!}{
\begin{tabular}{ccccccccc}
\toprule
\multicolumn{1}{l|}{\multirow{3}{*}{Method}} & \multicolumn{8}{c}{Various-LangSeg}                                                                                                                                                                                      \\ \cline{2-9} 
\multicolumn{1}{c|}{}                        & \multicolumn{2}{c|}{Explicit Semantic}                             & \multicolumn{2}{c|}{Generic Object}                             & \multicolumn{2}{c|}{Reasoning-Guided}                             & \multicolumn{2}{c}{Overall}                         \\
\multicolumn{1}{l|}{}                        & \multicolumn{1}{c}{gIoU} & \multicolumn{1}{c|}{cIoU} & \multicolumn{1}{c}{gIoU} & \multicolumn{1}{c|}{cIoU} & \multicolumn{1}{c}{gIoU} & \multicolumn{1}{c|}{cIoU} & \multicolumn{1}{c}{gIoU} & \multicolumn{1}{c}{cIoU} \\ \hline
\multicolumn{9}{c}{\small training-based methods}                                                                                                                                                                                                                              \\ \hline
\multicolumn{1}{l|}{ReLA*}                    &76.8                          &\multicolumn{1}{c|}{77.7 }     & 20.1                         & \multicolumn{1}{c|}{22.1}     &   25.2                       & \multicolumn{1}{c|}{21.2}     & 53.4                         & 53.8                         \\
\multicolumn{1}{l|}{OVSeg*}                   & 77.1                         & \multicolumn{1}{c|}{76.0}     &  23.2                        & \multicolumn{1}{c|}{23.0}     &        25.0                  & \multicolumn{1}{c|}{22.1}     & 54.4                         &  53.2                        \\
\multicolumn{1}{l|}{LISA-7B}                    &      81.9                    & \multicolumn{1}{c|}{83.1}     &   32.4                       & \multicolumn{1}{c|}{30.8}     &  46.8                        & \multicolumn{1}{c|}{36.7}     &   63.2                       &   64.2                       \\
\multicolumn{1}{l|}{LISA-13B}                    &   \textbf{82.8}                       & \multicolumn{1}{c|}{\textbf{83.9}}     &  35.3                        & \multicolumn{1}{c|}{40.3}     &   54.4                       & \multicolumn{1}{c|}{47.9}     & 65.8                         &        67.9                  \\
\multicolumn{1}{l|}{PixelLLM-7B}                    &  81.5                        & \multicolumn{1}{c|}{83.5}     & 31.2                         & \multicolumn{1}{c|}{31.5}     &  45.2                        & \multicolumn{1}{c|}{40.1}     & 62.3                         &    62.7                      \\
\multicolumn{1}{l|}{SAM4MLLM}                    & 82.1                         & \multicolumn{1}{c|}{83.5}     &  32.0                        & \multicolumn{1}{c|}{31.8}     &  45.2                        & \multicolumn{1}{c|}{37.1}     & 62.9                         &  62.3                        \\
\multicolumn{1}{l|}{Seg-Zero-7B}                &  81.8                        & \multicolumn{1}{c|}{81.0}     &  \textbf{41.4}                        & \multicolumn{1}{c|}{\textbf{43.5}}     &  \textbf{74.5}                        & \multicolumn{1}{c|}{\textbf{67.1}}     &    \textbf{70.0  }                    & \textbf{ 68.9 }                       \\\hline
\multicolumn{9}{c}{\small training-free methods}                                                                                                                                                                                                                               \\ \hline
\multicolumn{1}{l|}{Qwen2.5-VL-3B + SAM2-L}               &  80.1                        & \multicolumn{1}{c|}{77.4}     & 33.3                        & \multicolumn{1}{c|}{33.7}     & 61.0                        & \multicolumn{1}{c|}{49.5}     &  64.7                        &        61.3                  \\
\multicolumn{1}{l|}{Qwen2.5-VL-7B + SAM2-L}               &   80.8                       & \multicolumn{1}{c|}{79.7}     & 39.5                         & \multicolumn{1}{c|}{38.3}     &  70.1                        & \multicolumn{1}{c|}{58.9}     &  68.2                        & 65.4                         \\

\multicolumn{1}{l|}{Seg-Agent-3B (Ours)}               &   82.3                       & \multicolumn{1}{c|}{81.5}     &  40.8                        & \multicolumn{1}{c|}{31.5}     &    66.1                      & \multicolumn{1}{c|}{62.6}     &   68.8                       &    60.9                      \\
\multicolumn{1}{l|}{Seg-Agent-7B (Ours)}               & \textbf{83.0 }                        & \multicolumn{1}{c|}{\textbf{83.7}}     &  \textbf{41.0  }                      & \multicolumn{1}{c|}{\textbf{42.1}}     &    \textbf{75.2   }                   & \multicolumn{1}{c|}{\textbf{66.7} }    &  \textbf{70.6   }                     &     \textbf{68.5  }                   \\ 
\bottomrule
\end{tabular}}
\label{tab:various}
\end{table*}

%% file: table/ablation.tex
\begin{table}[t]
\centering
\caption{Ablation study on each module. GM, SM and RM denote generation module, selection module and refinement module, respectively.  We compare gIoU here. }
\vspace{-2mm}
\resizebox{0.9\linewidth}{!}{
\begin{tabular}{ccc|cccc}
\toprule
\multirow{2}{*}{GM} & \multirow{2}{*}{SM} & \multirow{2}{*}{RM} & \multicolumn{4}{c}{Various-LangSeg} \\
                    &                     &                     & ESS    & GOS    & RGS   & Overall   \\ \hline
\ding{55}                 & \ding{55}                 & \ding{55}                   &  80.8      & 39.5       & 70.1      & 68.2          \\
        \ding{51}                 & \ding{51}                 & \ding{55}                   &81.2        &  39.8      & 72.0      &    68.8       \\
            \ding{51}                 & \ding{55}                 & \ding{51}                   & 81.5       & 40.0       &  71.8     &  69.0         \\
 \ding{51}                    &  \ding{51}                    &    \ding{51}                  &  \textbf{83.0}      &  \textbf{41.0}      &   \textbf{75.2}    &   \textbf{70.6}        \\ 
 \bottomrule
\end{tabular}}

\label{tab:ablation}
\end{table}

%% file: table/setting.tex
\begin{table}[t]
\centering
\caption{Ablation study on base models. a, b, and c represent different combinations of MLLMs and base segmentation models, directly using a single-step reasoning approach. Seg-Agent (x) denotes using the configuration described in x to replace the corresponding part of Seg-Agent. We compare gIoU in this table.}
\vspace{-2mm}
\resizebox{\linewidth}{!}{
\begin{tabular}{l|cccc}
\toprule
\multicolumn{1}{l|}{\multirow{2}{*}{Setting}} & \multicolumn{4}{c}{Various-LangSeg} \\
\multicolumn{1}{l|}{}                         & ESS    & GOS    & RGS   & Overall   \\ \hline
a: InternVL3-8B + SAM2-L                                         &  79.2      & 35.3      & 67.1     & 65.7    \\
b: Qwen2.5-VL-7B + SAM-L                                    & 80.1       &  37.5      & 69.9      & 67.3          \\
c: Qwen2.5-VL-7B + SAM2-L                              &  80.8      &  39.5      & 70.1      & 68.2          \\ \hline
Seg-Agent (a)                             & 79.8             & 36.3      &    67.5  &  66.4     \\ 
Seg-Agent (b)                             &  81.1              & 39.5      &     71.2    &68.6  \\
Seg-Agent (c)                             & 83.0       &  41.0      &  75.2     &  70.6         \\
\bottomrule
\end{tabular}}
\vspace{-4mm}

\label{tab:setting}
\end{table}

%% file: reference.bib
@string(CVPR  = "{IEEE Conf. Comput. Vis. Pattern Recog.}")

@inproceedings{lisa,
  title={Lisa: Reasoning segmentation via large language model},
  author={Lai, Xin and Tian, Zhuotao and Chen, Yukang and Li, Yanwei and Yuan, Yuhui and Liu, Shu and Jia, Jiaya},
  booktitle={Proceedings of the IEEE/CVF Conference on Computer Vision and Pattern Recognition},
  pages={9579--9589},
  year={2024}
}

@article{segzero,
  title        = {Seg-Zero: Reasoning-Chain Guided  Segmentation via Cognitive Reinforcement},
  author       = {Liu, Yuqi and Peng, Bohao and Zhong, Zhisheng and Yue, Zihao and Lu, Fanbin and Yu, Bei and Jia, Jiaya},
  journal      = {arXiv preprint arXiv:2503.06520},
  year         = {2025}
}

@article{SoM,
  title={Set-of-mark prompting unleashes extraordinary visual grounding in gpt-4v},
  author={Yang, Jianwei and Zhang, Hao and Li, Feng and Zou, Xueyan and Li, Chunyuan and Gao, Jianfeng},
  journal={arXiv preprint arXiv:2310.11441},
  year={2023}
}

@article{groundedsam,
  title={Grounded sam: Assembling open-world models for diverse visual tasks},
  author={Ren, Tianhe and Liu, Shilong and Zeng, Ailing and Lin, Jing and Li, Kunchang and Cao, He and Chen, Jiayu and Huang, Xinyu and Chen, Yukang and Yan, Feng and others},
  journal={arXiv preprint arXiv:2401.14159},
  year={2024}
}

@article{thinkbeforesegment,
  title={Think before you segment: High-quality reasoning segmentation with gpt chain of thoughts},
  author={Kao, Shiu-hong and Tai, Yu-Wing and Tang, Chi-Keung},
  journal={arXiv preprint arXiv:2503.07503},
  year={2025}
}

@inproceedings{SAM,
  title={Segment anything},
  author={Kirillov, Alexander and Mintun, Eric and Ravi, Nikhila and Mao, Hanzi and Rolland, Chloe and Gustafson, Laura and Xiao, Tete and Whitehead, Spencer and Berg, Alexander C and Lo, Wan-Yen and others},
  booktitle={Proceedings of the IEEE/CVF international conference on computer vision},
  pages={4015--4026},
  year={2023}
}

@article{SAM2,
  title={Sam 2: Segment anything in images and videos},
  author={Ravi, Nikhila and Gabeur, Valentin and Hu, Yuan-Ting and Hu, Ronghang and Ryali, Chaitanya and Ma, Tengyu and Khedr, Haitham and R{\"a}dle, Roman and Rolland, Chloe and Gustafson, Laura and others},
  journal={arXiv preprint arXiv:2408.00714},
  year={2024}
}

@inproceedings{sam4mllm,
  title={Sam4mllm: Enhance multi-modal large language model for referring expression segmentation},
  author={Chen, Yi-Chia and Li, Wei-Hua and Sun, Cheng and Wang, Yu-Chiang Frank and Chen, Chu-Song},
  booktitle={European Conference on Computer Vision},
  pages={323--340},
  year={2024},
  organization={Springer}
}

@inproceedings{pixellm,
  title={Pixellm: Pixel reasoning with large multimodal model},
  author={Ren, Zhongwei and Huang, Zhicheng and Wei, Yunchao and Zhao, Yao and Fu, Dongmei and Feng, Jiashi and Jin, Xiaojie},
  booktitle={Proceedings of the IEEE/CVF Conference on Computer Vision and Pattern Recognition},
  pages={26374--26383},
  year={2024}
}

@article{qwenvl2.5,
  title={Qwen2. 5-vl technical report},
  author={Bai, Shuai and Chen, Keqin and Liu, Xuejing and Wang, Jialin and Ge, Wenbin and Song, Sibo and Dang, Kai and Wang, Peng and Wang, Shijie and Tang, Jun and others},
  journal={arXiv preprint arXiv:2502.13923},
  year={2025}
}

@article{gpt4,
  title={Gpt-4 technical report},
  author={Achiam, Josh and Adler, Steven and Agarwal, Sandhini and Ahmad, Lama and Akkaya, Ilge and Aleman, Florencia Leoni and Almeida, Diogo and Altenschmidt, Janko and Altman, Sam and Anadkat, Shyamal and others},
  journal={arXiv preprint arXiv:2303.08774},
  year={2023}
}

@article{SENet,
  title={A simple yet effective network based on vision transformer for camouflaged object and salient object detection},
  author={Hao, Chao and Yu, Zitong and Liu, Xin and Xu, Jun and Yue, Huanjing and Yang, Jingyu},
  journal={IEEE Transactions on Image Processing},
  year={2025},
  publisher={IEEE}
}

@misc{JoNet,
      title={Distribution-Specific Learning for Joint Salient and Camouflaged Object Detection}, 
      author={Chao Hao and Zitong Yu and Xin Liu and Yuhao Wang and Weicheng Xie and Jingang Shi and Huanjing Yue and Jingyu Yang},
      year={2025},
      eprint={2508.06063},
      archivePrefix={arXiv},
      primaryClass={cs.CV},
      url={https://arxiv.org/abs/2508.06063}, 
}

@article{llava,
  title={Visual instruction tuning},
  author={Liu, Haotian and Li, Chunyuan and Wu, Qingyang and Lee, Yong Jae},
  journal={Advances in neural information processing systems},
  volume={36},
  pages={34892--34916},
  year={2023}
}

@article{omg-llava,
  title={Omg-llava: Bridging image-level, object-level, pixel-level reasoning and understanding},
  author={Zhang, Tao and Li, Xiangtai and Fei, Hao and Yuan, Haobo and Wu, Shengqiong and Ji, Shunping and Loy, Chen Change and Yan, Shuicheng},
  journal={Advances in neural information processing systems},
  volume={37},
  pages={71737--71767},
  year={2024}
}

@article{sa2va,
  title={Sa2va: Marrying sam2 with llava for dense grounded understanding of images and videos},
  author={Yuan, Haobo and Li, Xiangtai and Zhang, Tao and Huang, Zilong and Xu, Shilin and Ji, Shunping and Tong, Yunhai and Qi, Lu and Feng, Jiashi and Yang, Ming-Hsuan},
  journal={arXiv preprint arXiv:2501.04001},
  year={2025}
}

@inproceedings{mask2former,
  title={Masked-attention Mask Transformer for Universal Image Segmentation},
  author={Bowen Cheng and Ishan Misra and Alexander G. Schwing and Alexander Kirillov and Rohit Girdhar},
  journal={CVPR},
  year={2022}
}

@inproceedings{segformer,
  title={SegFormer: Simple and Efficient Design for Semantic Segmentation with Transformers},
  author={Xie, Enze and Wang, Wenhai and Yu, Zhiding and Anandkumar, Anima and Alvarez, Jose M and Luo, Ping},
  booktitle={Neural Information Processing Systems (NeurIPS)},
  year={2021}
}

@inproceedings{SETR,
    title={Rethinking Semantic Segmentation from a Sequence-to-Sequence Perspective with Transformers}, 
    author={Zheng, Sixiao and Lu, Jiachen and Zhao, Hengshuang and Zhu, Xiatian and Luo, Zekun and Wang, Yabiao and Fu, Yanwei and Feng, Jianfeng and Xiang, Tao and Torr, Philip H.S. and Zhang, Li},
    booktitle={CVPR},
    year={2021}
}

@inproceedings{refcoco,
    title = "{R}efer{I}t{G}ame: Referring to Objects in Photographs of Natural Scenes",
    author = "Kazemzadeh, Sahar  and
      Ordonez, Vicente  and
      Matten, Mark  and
      Berg, Tamara",
    editor = "Moschitti, Alessandro  and
      Pang, Bo  and
      Daelemans, Walter",
    booktitle = "Proceedings of the 2014 Conference on Empirical Methods in Natural Language Processing ({EMNLP})",
    month = oct,
    year = "2014",
    address = "Doha, Qatar",
    publisher = "Association for Computational Linguistics",
    url = "https://aclanthology.org/D14-1086/",
    doi = "10.3115/v1/D14-1086",
    pages = "787--798"
}

@inproceedings{refcocog,
  title = {Generation and Comprehension of Unambiguous Object Descriptions},
  url = {http://dx.doi.org/10.1109/CVPR.2016.9},
  DOI = {10.1109/cvpr.2016.9},
  booktitle = {2016 IEEE Conference on Computer Vision and Pattern Recognition (CVPR)},
  publisher = {IEEE},
  author = {Mao,  Junhua and Huang,  Jonathan and Toshev,  Alexander and Camburu,  Oana and Yuille,  Alan and Murphy,  Kevin},
  year = {2016},
  month = jun,
  pages = {11–20}
}

@InProceedings{GSVA,
    author    = {Xia, Zhuofan and Han, Dongchen and Han, Yizeng and Pan, Xuran and Song, Shiji and Huang, Gao},
    title     = {GSVA: Generalized Segmentation via Multimodal Large Language Models},
    booktitle = {Proceedings of the IEEE/CVF Conference on Computer Vision and Pattern Recognition (CVPR)},
    month     = {June},
    year      = {2024},
    pages     = {3858-3869}
}

@inproceedings{maskformer,
  title={Per-Pixel Classification is Not All You Need for Semantic Segmentation},
  author={Bowen Cheng and Alexander G. Schwing and Alexander Kirillov},
  journal={NeurIPS},
  year={2021}
}

@misc{androidworld,
      title={AndroidWorld: A Dynamic Benchmarking Environment for Autonomous Agents},
      author={Christopher Rawles and Sarah Clinckemaillie and Yifan Chang and Jonathan Waltz and Gabrielle Lau and Marybeth Fair and Alice Li and William Bishop and Wei Li and Folawiyo Campbell-Ajala and Daniel Toyama and Robert Berry and Divya Tyamagundlu and Timothy Lillicrap and Oriana Riva},
      year={2024},
      eprint={2405.14573},
      archivePrefix={arXiv},
      primaryClass={cs.AI},
      url={https://arxiv.org/abs/2405.14573},
}

@misc{guigrounding,
      title={SeeClick: Harnessing GUI Grounding for Advanced Visual GUI Agents}, 
      author={Kanzhi Cheng and Qiushi Sun and Yougang Chu and Fangzhi Xu and Yantao Li and Jianbing Zhang and Zhiyong Wu},
      year={2024},
      eprint={2401.10935},
      archivePrefix={arXiv},
      primaryClass={cs.HC},
      url={https://arxiv.org/abs/2401.10935}, 
}

@misc{vqa,
      title={VQA: Visual Question Answering}, 
      author={Aishwarya Agrawal and Jiasen Lu and Stanislaw Antol and Margaret Mitchell and C. Lawrence Zitnick and Dhruv Batra and Devi Parikh},
      year={2016},
      eprint={1505.00468},
      archivePrefix={arXiv},
      primaryClass={cs.CL},
      url={https://arxiv.org/abs/1505.00468}, 
}

@article{captioning,
   title={Deep Learning Approaches on Image Captioning: A Review},
   volume={56},
   ISSN={1557-7341},
   url={http://dx.doi.org/10.1145/3617592},
   DOI={10.1145/3617592},
   number={3},
   journal={ACM Computing Surveys},
   publisher={Association for Computing Machinery (ACM)},
   author={Ghandi, Taraneh and Pourreza, Hamidreza and Mahyar, Hamidreza},
   year={2023},
   month=oct, pages={1–39} }

@article{
zhang2024multimodal,
title={Multimodal Chain-of-Thought Reasoning in Language Models},
author={Zhuosheng Zhang and Aston Zhang and Mu Li and hai zhao and George Karypis and Alex Smola},
journal={Transactions on Machine Learning Research},
issn={2835-8856},
year={2024},
url={https://openreview.net/forum?id=y1pPWFVfvR},
note={}
}

@misc{visionllmv2,
      title={VisionLLM v2: An End-to-End Generalist Multimodal Large Language Model for Hundreds of Vision-Language Tasks}, 
      author={Jiannan Wu and Muyan Zhong and Sen Xing and Zeqiang Lai and Zhaoyang Liu and Zhe Chen and Wenhai Wang and Xizhou Zhu and Lewei Lu and Tong Lu and Ping Luo and Yu Qiao and Jifeng Dai},
      year={2024},
      eprint={2406.08394},
      archivePrefix={arXiv},
      primaryClass={cs.CV},
      url={https://arxiv.org/abs/2406.08394}, 
}

@article{bert,
  title={BERT: Pre-training of Deep Bidirectional Transformers for Language Understanding},
  author={Devlin, Jacob and Chang, Ming-Wei and Lee, Kenton and Toutanova, Kristina},
  journal={arXiv preprint arXiv:1810.04805},
  year={2018}
}

@misc{clip,
      title={Learning Transferable Visual Models From Natural Language Supervision}, 
      author={Alec Radford and Jong Wook Kim and Chris Hallacy and Aditya Ramesh and Gabriel Goh and Sandhini Agarwal and Girish Sastry and Amanda Askell and Pamela Mishkin and Jack Clark and Gretchen Krueger and Ilya Sutskever},
      year={2021},
      eprint={2103.00020},
      archivePrefix={arXiv},
      primaryClass={cs.CV},
      url={https://arxiv.org/abs/2103.00020}, 
}

@article{groundingdino,
  title={Grounding dino: Marrying dino with grounded pre-training for open-set object detection},
  author={Liu, Shilong and Zeng, Zhaoyang and Ren, Tianhe and Li, Feng and Zhang, Hao and Yang, Jie and Li, Chunyuan and Yang, Jianwei and Su, Hang and Zhu, Jun and others},
  journal={arXiv preprint arXiv:2303.05499},
  year={2023}
}

@misc{LAVT,
      title={LAVT: Language-Aware Vision Transformer for Referring Image Segmentation}, 
      author={Zhao Yang and Jiaqi Wang and Yansong Tang and Kai Chen and Hengshuang Zhao and Philip H. S. Torr},
      year={2022},
      eprint={2112.02244},
      archivePrefix={arXiv},
      primaryClass={cs.CV},
      url={https://arxiv.org/abs/2112.02244}, 
}

@misc{GRES,
      title={GRES: Generalized Referring Expression Segmentation}, 
      author={Chang Liu and Henghui Ding and Xudong Jiang},
      year={2023},
      eprint={2306.00968},
      archivePrefix={arXiv},
      primaryClass={cs.CV},
      url={https://arxiv.org/abs/2306.00968}, 
}

@inproceedings{NC4K,
  title={Simultaneously localize, segment and rank the camouflaged objects},
  author={Lv, Yunqiu and Zhang, Jing and Dai, Yuchao and Li, Aixuan and Liu, Bowen and Barnes, Nick and Fan, Deng-Ping},
  booktitle={Proceedings of the IEEE/CVF Conference on Computer Vision and Pattern Recognition},
  pages={11591--11601},
  year={2021}
}

@inproceedings{PASCAL-S,
  title={The secrets of salient object segmentation},
  author={Li, Yin and Hou, Xiaodi and Koch, Christof and Rehg, James M and Yuille, Alan L},
  booktitle={Proceedings of the IEEE conference on computer vision and pattern recognition},
  pages={280--287},
  year={2014}
}

@inproceedings{SBU,
  title={Large-scale training of shadow detectors with noisily-annotated shadow examples},
  author={Vicente, Tom{\'a}s F Yago and Hou, Le and Yu, Chen-Ping and Hoai, Minh and Samaras, Dimitris},
  booktitle={Computer Vision--ECCV 2016: 14th European Conference, Amsterdam, The Netherlands, October 11-14, 2016, Proceedings, Part VI 14},
  pages={816--832},
  year={2016},
  organization={Springer}
}

@INPROCEEDINGS{casia,
  author={Dong, Jing and Wang, Wei and Tan, Tieniu},
  booktitle={2013 IEEE China Summit and International Conference on Signal and Information Processing}, 
  title={CASIA Image Tampering Detection Evaluation Database}, 
  year={2013},
  volume={},
  number={},
  pages={422-426},
  doi={10.1109/ChinaSIP.2013.6625374}}

@misc{coco,
      title={Microsoft COCO: Common Objects in Context}, 
      author={Tsung-Yi Lin and Michael Maire and Serge Belongie and Lubomir Bourdev and Ross Girshick and James Hays and Pietro Perona and Deva Ramanan and C. Lawrence Zitnick and Piotr Dollár},
      year={2015},
      eprint={1405.0312},
      archivePrefix={arXiv},
      primaryClass={cs.CV},
      url={https://arxiv.org/abs/1405.0312}, 
}

@inproceedings{cris,
  title={CRIS: CLIP-Driven Referring Image Segmentation},
  author={Wang, Zhaoqing and Lu, Yu and Li, Qiang and Tao, Xunqiang and Guo, Yandong and Gong, Mingming and Liu, Tongliang},
  booktitle={Proceedings of the IEEE/CVF conference on computer vision and pattern recognition},
  year={2022}
}

@misc{text4seg,
      title={Text4Seg: Reimagining Image Segmentation as Text Generation}, 
      author={Mengcheng Lan and Chaofeng Chen and Yue Zhou and Jiaxing Xu and Yiping Ke and Xinjiang Wang and Litong Feng and Wayne Zhang},
      year={2024},
      eprint={2410.09855},
      archivePrefix={arXiv},
      primaryClass={cs.CV},
      url={https://arxiv.org/abs/2410.09855}, 
}

@InProceedings{PerceptionGPT,
    author    = {Pi, Renjie and Yao, Lewei and Gao, Jiahui and Zhang, Jipeng and Zhang, Tong},
    title     = {PerceptionGPT: Effectively Fusing Visual Perception into LLM},
    booktitle = {Proceedings of the IEEE/CVF Conference on Computer Vision and Pattern Recognition (CVPR)},
    month     = {June},
    year      = {2024},
    pages     = {27124-27133}
}

@inproceedings{ovseg,
  title={Open-vocabulary semantic segmentation with mask-adapted clip},
  author={Liang, Feng and Wu, Bichen and Dai, Xiaoliang and Li, Kunpeng and Zhao, Yinan and Zhang, Hang and Zhang, Peizhao and Vajda, Peter and Marculescu, Diana},
  booktitle={Proceedings of the IEEE/CVF Conference on Computer Vision and Pattern Recognition},
  pages={7061--7070},
  year={2023}
}

@misc{xdecoder,
      title={Generalized Decoding for Pixel, Image, and Language}, 
      author={Xueyan Zou and Zi-Yi Dou and Jianwei Yang and Zhe Gan and Linjie Li and Chunyuan Li and Xiyang Dai and Harkirat Behl and Jianfeng Wang and Lu Yuan and Nanyun Peng and Lijuan Wang and Yong Jae Lee and Jianfeng Gao},
      year={2022},
      eprint={2212.11270},
      archivePrefix={arXiv},
      primaryClass={cs.CV},
      url={https://arxiv.org/abs/2212.11270}, 
}

@misc{SEEM,
      title={Segment Everything Everywhere All at Once}, 
      author={Xueyan Zou and Jianwei Yang and Hao Zhang and Feng Li and Linjie Li and Jianfeng Wang and Lijuan Wang and Jianfeng Gao and Yong Jae Lee},
      year={2023},
      eprint={2304.06718},
      archivePrefix={arXiv},
      primaryClass={cs.CV},
      url={https://arxiv.org/abs/2304.06718}, 
}

@misc{grpo,
      title={DeepSeekMath: Pushing the Limits of Mathematical Reasoning in Open Language Models}, 
      author={Zhihong Shao and Peiyi Wang and Qihao Zhu and Runxin Xu and Junxiao Song and Xiao Bi and Haowei Zhang and Mingchuan Zhang and Y. K. Li and Y. Wu and Daya Guo},
      year={2024},
      eprint={2402.03300},
      archivePrefix={arXiv},
      primaryClass={cs.CL},
      url={https://arxiv.org/abs/2402.03300}, 
}

@article{sod,
  title={Salient object detection: A benchmark},
  author={Borji, Ali and Cheng, Ming-Ming and Jiang, Huaizu and Li, Jia},
  journal={IEEE transactions on image processing},
  volume={24},
  number={12},
  pages={5706--5722},
  year={2015},
  publisher={IEEE}
}

@inproceedings{cod,
  title={Camouflaged object detection},
  author={Fan, Deng-Ping and Ji, Ge-Peng and Sun, Guolei and Cheng, Ming-Ming and Shen, Jianbing and Shao, Ling},
  booktitle={Proceedings of the IEEE/CVF conference on computer vision and pattern recognition},
  pages={2777--2787},
  year={2020}
}

@misc{internvl3,
      title={InternVL3: Exploring Advanced Training and Test-Time Recipes for Open-Source Multimodal Models}, 
      author={Jinguo Zhu and Weiyun Wang and Zhe Chen and Zhaoyang Liu and Shenglong Ye and Lixin Gu and Hao Tian and Yuchen Duan and Weijie Su and Jie Shao and Zhangwei Gao and Erfei Cui and Xuehui Wang and Yue Cao and Yangzhou Liu and Xingguang Wei and Hongjie Zhang and Haomin Wang and Weiye Xu and Hao Li and Jiahao Wang and Nianchen Deng and Songze Li and Yinan He and Tan Jiang and Jiapeng Luo and Yi Wang and Conghui He and Botian Shi and Xingcheng Zhang and Wenqi Shao and Junjun He and Yingtong Xiong and Wenwen Qu and Peng Sun and Penglong Jiao and Han Lv and Lijun Wu and Kaipeng Zhang and Huipeng Deng and Jiaye Ge and Kai Chen and Limin Wang and Min Dou and Lewei Lu and Xizhou Zhu and Tong Lu and Dahua Lin and Yu Qiao and Jifeng Dai and Wenhai Wang},
      year={2025},
      eprint={2504.10479},
      archivePrefix={arXiv},
      primaryClass={cs.CV},
      url={https://arxiv.org/abs/2504.10479}, 
}

@inproceedings{mmr,
title={{MMR}: A Large-scale Benchmark Dataset for Multi-target and Multi-granularity Reasoning Segmentation},
author={Donggon Jang and Yucheol Cho and Suin Lee and Taehyeon Kim and Daeshik Kim},
booktitle={The Thirteenth International Conference on Learning Representations},
year={2025},
url={https://openreview.net/forum?id=mzL19kKE3r}
}

@inproceedings{segllm,
title={Seg{LLM}: Multi-round Reasoning Segmentation with Large Language Models},
author={XuDong Wang and Shaolun Zhang and Shufan Li and Kehan Li and Konstantinos Kallidromitis and Yusuke Kato and Kazuki Kozuka and Trevor Darrell},
booktitle={The Thirteenth International Conference on Learning Representations},
year={2025},
url={https://openreview.net/forum?id=Pm1NXHgzyf}
}

@article{deepseekr1,
  title={Deepseek-r1: Incentivizing reasoning capability in llms via reinforcement learning},
  author={Guo, Daya and Yang, Dejian and Zhang, Haowei and Song, Junxiao and Zhang, Ruoyu and Xu, Runxin and Zhu, Qihao and Ma, Shirong and Wang, Peiyi and Bi, Xiao and others},
  journal={arXiv preprint arXiv:2501.12948},
  year={2025}
}

@misc{yuv20k,
      title={YUV20K: A Complexity-Driven Benchmark and Trajectory-Aware Alignment Model for Video Camouflaged Object Detection}, 
      author={Yiyu Liu and Shuo Ye and Chao Hao and Zitong Yu},
      year={2026},
      eprint={2604.09985},
      archivePrefix={arXiv},
      primaryClass={cs.CV},
      url={https://arxiv.org/abs/2604.09985}, 
}
